\documentclass[runningheads]{llncs}
\usepackage{graphicx}
\usepackage{subfig}
\usepackage{amsmath,amssymb} 
\usepackage{color}
\usepackage[width=122mm,left=12mm,paperwidth=146mm,height=193mm,top=12mm,paperheight=217mm]{geometry}
\usepackage{spverbatim}
\usepackage{etoolbox}
\makeatletter
\preto{\@verbatim}{\topsep=1pt \partopsep=1pt}
\makeatother

\usepackage{array}
\newcolumntype{L}[1]{>{\raggedright\let\newline\\\arraybackslash\hspace{0pt}}m{#1}}
\newcolumntype{C}[1]{>{\centering\let\newline\\\arraybackslash\hspace{0pt}}m{#1}}
\newcolumntype{R}[1]{>{\raggedleft\let\newline\\\arraybackslash\hspace{0pt}}m{#1}}

\begin{document}
\pagestyle{headings}
\mainmatter

\title{Geometric Neural Phrase Pooling: \\ Modeling the Spatial Co-occurrence of Neurons} 

\titlerunning{Geometric Neural Phrase Pooling}

\authorrunning{L. Xie, Q. Tian, J. Flynn, J. Wang and A. Yuille}

\author{Lingxi Xie\textsuperscript{1}, Qi Tian\textsuperscript{2},
John Flynn\textsuperscript{3}, Jingdong Wang\textsuperscript{4}, Alan Yuille\textsuperscript{5}}

\institute{
\textsuperscript{1,5}Center for Imaging Science, The Johns Hopkins University, Baltimore, MD, USA\\
\textsuperscript{2}Department of Computer Science, University of Texas at San Antonio, TX, USA\\
\textsuperscript{3}Department of Statistics, University of California, Los Angeles, CA, USA\\
\textsuperscript{4}Microsoft Research, Beijing, China\\
\textsuperscript{1}{\tt\small 198808xc@gmail.com}\quad
\textsuperscript{2}{\tt\small qitian@cs.utsa.edu}\\
\textsuperscript{3}{\tt\small john\_flynn@mac.com}\quad
\textsuperscript{4}{\tt\small jingdw@microsoft.com}\quad
\textsuperscript{5}{\tt\small alan.l.yuille@gmail.com}\\
\textcolor{red}{Codes: {\tt http://bigml.cs.tsinghua.edu.cn/\~{}lingxi/Projects/GNPP.html}}
}

\maketitle

\begin{abstract}
Deep Convolutional Neural Networks (CNNs) are playing important roles in state-of-the-art visual recognition.
This paper focuses on modeling the spatial co-occurrence of neuron responses, which is less studied in the previous work.
For this, we consider the neurons in the hidden layer as {\em neural words},
and construct a set of {\em geometric neural phrases} on top of them.
The idea that grouping neural words into neural phrases is borrowed from the Bag-of-Visual-Words (BoVW) model.
Next, the {\bf Geometric Neural Phrase Pooling} (GNPP) algorithm is proposed to efficiently encode these neural phrases.
GNPP acts as a new type of hidden layer, which punishes the isolated neuron responses after convolution,
and can be inserted into a CNN model with little extra computational overhead.
Experimental results show that GNPP produces significant and consistent accuracy gain in image classification.
\end{abstract}

\section{Introduction}
\label{Introduction}

We have witnessed a significant revolution in computer vision brought by the deep Convolutional Neural Networks (CNNs).
With powerful computational resources ({\em e.g.}, GPUs)
and a large amount of labeled training data ({\em e.g.}, \cite{Deng_2009_ImageNet}),
a hierarchical structure containing different levels of visual concepts is constructed and trained~\cite{Krizhevsky_2012_ImageNet}
to produce impressive performance on large-scale visual recognition tasks~\cite{Russakovsky_2015_ImageNet}.
A pre-trained deep network is also capable of generating deep features for various tasks,
such as image classification~\cite{Jia_2014_CAFFE}\cite{Donahue_2014_DeCAF},
image retrieval~\cite{Razavian_2014_CNN}\cite{Xie_2015_Image} and object detection~\cite{Girshick_2014_Rich}\cite{Girshick_2015_Fast}.

CNN is composed of several stacked layers, each of which contains a number of neurons.
We argue that modeling the co-occurrence of neuron responses is important, whereas less studied in the previous work.
For this, we define a set of {\em geometric neural phrases} on the basis of the hidden neurons,
and propose the {\bf Geometric Neural Phrase Pooling} (GNPP) algorithm to encode them efficiently.
GNPP can be regarded as a new type of layer, and inserted into a network with little computational overhead
({\em e.g.}, $1.29\%$ and $2.52\%$ extra time and memory costs in the experiments on {\bf ImageNet}).
We explain the behavior of GNPP by noting that it punishes the isolated neuron responses,
and that the isolated responses are often less reliable than clustered ones, especially in the high-level network layers.
Experimental results show that adding GNPP layers boosts image classification accuracy significantly and consistently.
Later, we will discuss the benefits brought by the GNPP layer from different points of view,
showing that GNPP produces better internal representation, builds latent connections, and accelerates the network training process.

The remainder of this paper is organized as follows.
Section~\ref{RelatedWork} briefly introduces related work.
Section~\ref{GNPP} introduces the GNPP layer, and Section~\ref{Experiments} shows experimental results.
We discuss the benefits brought by adding GNPP layers in Section~\ref{Benefits}.
Finally, we conclude this work in Section~\ref{Conclusions}.

\section{Related Work}
\label{RelatedWork}

\subsection{The Bag-of-Visual-Words Model}
\label{RelatedWork:BoVW}

The Bag-of-Visual-Words (BoVW) model~\cite{Csurka_2004_Visual} represents each image as a high-dimensional vector.
It typically consists of three stages, {\em i.e.}, descriptor extraction, feature encoding and feature aggregation.

Due to the limited descriptive ability of raw pixels,
handcrafted descriptors such as SIFT~\cite{Lowe_2004_Distinctive},
HOG~\cite{Dalal_2005_Histograms} or other variants~\cite{Xie_2015_RIDE} are extracted.
The set of local descriptors on an image is denoted as
${\mathcal{D}}={\left\{\left(\mathbf{d}_m,\mathbf{l}_m\right)\right\}_{m=1}^M}$,
where $M$ is the number of descriptors,
$\mathbf{d}_m$ is the description vector and $\mathbf{l}_m$ is the 2D location of the $m$-th word.
A visual vocabulary or codebook is then built to capture the data distribution in feature space.
The codebook is a set of codewords: ${\mathcal{B}}={\left\{\mathbf{c}_b\right\}_{b=1}^B}$,
in which $B$ is the codebook size and each codeword has the same dimension with the descriptors.
Each descriptor $\mathbf{d}_m$ is then quantized onto the codebook as a visual word ${\mathbf{f}_m}\in{\mathbb{R}_{\geqslant0}^B}$.
Effective feature quantization algorithms include sparse coding~\cite{Yang_2009_Linear}\cite{Wang_2010_Locality}
and high-dimensional encoding~\cite{Perronnin_2010_Improving}\cite{Zhou_2010_Image}\cite{Kobayashi_2014_Dirichlet}.
${\mathcal{F}}={\left\{\left(\mathbf{f}_m,\mathbf{l}_m\right)\right\}_{m=1}^M}$ is the set of visual words.
Finally, these words are aggregated as an image-level representation vector~\cite{Lazebnik_2006_Beyond}\cite{Feng_2011_Geometric}.
These Image-level vectors are then normalized and fed into
machine learning algorithms~\cite{Fan_2008_LIBLINEAR} for training and testing,
or used in some training-free image classification algorithms~\cite{Boiman_2008_Defense}\cite{Xie_2015_Image}.

\subsection{Geometric Phrase Pooling}
\label{RelatedWork:GPP}

The basic unit in the BoVW model is a {\em visual word}, {\em i.e.}, a quantized local descriptor.
Dealing with individual visual words does not consider the spatial co-occurrence of visual features.
To this end, researchers propose {\em visual phrase}~\cite{Yuan_2007_Discovery}\cite{Zhang_2011_Image}
as a mid-level data structure connecting low-level descriptors and high-level visual concepts~\cite{Zhang_2009_Descriptive}.
A visual phrase is often defined as a group of neighboring visual words~\cite{Zhang_2009_Descriptive}\cite{Xie_2014_Spatial}.
It can be used to filter out the false matches in object retrieval~\cite{Zhang_2011_Image}\cite{Jiang_2012_Randomized},
or improve the descriptive ability of visual features for image classification~\cite{Xie_2014_Spatial}\cite{Xie_2013_Hierarchical}.

Geometric Phrase Pooling (GPP)~\cite{Xie_2014_Spatial} is an efficient algorithm for extracting and encoding visual phrases.
GPP starts from constructing, for each visual word, a {\em geometric visual phrase}, which is a group of visual words:
${\mathcal{G}_m}={\left(\mathbf{f}_m,\mathbf{l}_m\right)\cup
    \left\{\left(\mathbf{f}_m^{\left(k\right)},\mathbf{l}_m^{\left(k\right)}\right)\right\}_{k=1}^K}$.
In $\mathcal{G}_m$, $\left(\mathbf{f}_m,\mathbf{l}_m\right)$ is the {\em central word},
and all the other $K$ words are {\em side words},
located in a small neighborhood $\mathcal{N}_m$ of the central position $\mathbf{l}_m$.
GPP encodes each geometric visual phrase $\mathcal{G}_m$
by adding the maximal response of the side words to the central word:
${\mathbf{p}_m}={\mathbf{f}_m+\max_{k=1}^{K}\left\{s_m^{\left(k\right)}\times\mathbf{f}_m^{\left(k\right)}\right\}}$,
where $\max_{k=1}^{K}\left\{\cdot\right\}$ denotes dimension-wise maximization.
Note that the central word is not included in the maximization.
$s_m^{\left(k\right)}$ is the smoothing weight of the $k$-th side word in $\mathcal{G}_m$.
Most often, $s_m^{\left(k\right)}$ is determined by the Euclidean distance between $\mathbf{l}_m^{\left(k\right)}$ and $\mathbf{l}_m$,
{\em e.g.}, ${s_m^{\left(k\right)}}=
    {\exp\!\left\{-\tau\times\left\|\mathbf{l}_m-\mathbf{l}_m^{\left(k\right)}\right\|_2\right\}}$,
where ${\tau}>{0}$ is the pre-defined smoothing parameter.
Note that, at least in theory, the GPP algorithm can be applied to any data with a spatial attribute.

\subsection{Convolutional Neural Networks}
\label{RelatedWork:CNN}

The Convolutional Neural Network (CNN) serves as a hierarchical model for large-scale visual recognition.
It is based on the observation that a network with enough neurons is able to fit any complicated data distribution.
In past years, neural networks were shown effective for simple recognition tasks~\cite{LeCun_1990_Handwritten}.
More recently, the availability of large-scale training data ({\em e.g.}, ImageNet~\cite{Deng_2009_ImageNet}) and powerful GPUs
make it possible to train deep CNNs~\cite{Krizhevsky_2012_ImageNet} which significantly outperform BoVW models.
A CNN is composed of several stacked layers.
In each of them, responses from the previous layer are convoluted with a filter bank and activated by a differentiable non-linearity.
Hence, a CNN can be considered as a composite function,
which is trained by back-propagating error signals defined by the difference between supervision and prediction at the top layer.
Efficient methods were proposed to help CNNs converge faster and prevent over-fitting,
such as ReLU activation~\cite{Krizhevsky_2012_ImageNet},
batch normalization~\cite{Ioffe_2015_Batch} and regularization~\cite{Hinton_2012_Improving}\cite{Xie_2016_DisturbLabel}.
It is believed that deeper networks
produce better recognition results~\cite{Simonyan_2015_Very}\cite{Szegedy_2015_Going}\cite{He_2015_Deep}.

The intermediate responses of CNNs, {\em i.e.}, the so-called deep features,
serve as effective image descriptions~\cite{Donahue_2014_DeCAF}, or a set of latent visual attributes~\cite{Zhang_2014_PANDA}.
They can be used for various types of vision tasks,
including image classification~\cite{Jia_2014_CAFFE}\cite{Xie_2016_InterActive},
image retrieval~\cite{Razavian_2014_CNN}\cite{Xie_2015_Image} and object detection~\cite{Girshick_2014_Rich}.
A discussion of how different CNN configurations impact deep feature performance is available in~\cite{Chatfield_2014_Return}.

\section{Geometric Neural Phrase Pooling}
\label{GNPP}

This section presents the Geometric Neural Phrase Pooling (GNPP) algorithm and its application to improve the CNN model.

\subsection{The GNPP Layer}
\label{GNPP:Layer}

We start with a hidden layer $\mathbf{X}$ in the CNN model.
$\mathbf{X}$ is a 3D neuron cube with $W\times H\times D$ neurons, where $W$, $H$ and $D$ are the width, height and depth of the cube.
The response of each neuron corresponds to
the inner-product of a local patch in the previous layer and a filter (convolutional kernel).
We naturally consider the data as a set of $D$-dimensional {\em visual words} indexed over a 2D spatial domain.
We denote the set as ${\mathcal{X}}={\left\{\mathbf{x}_{w,h}\right\}_{w=1,h=1}^{W,H}}$,
in which ${\mathbf{x}_{w,h}}\in{\mathbb{R}_{\geqslant0}^{D}}$ for each $w$ and $h$.
The spatial domain coordinate $\left(w,h\right)$ is not the same as the pixel coordinate $\left(a,b\right)$ in the original image,
but they are linearly corresponded.

\newcommand{\figurewidth}{12.0cm}
\begin{figure}[t]
\begin{center}
    \includegraphics[width=\figurewidth]{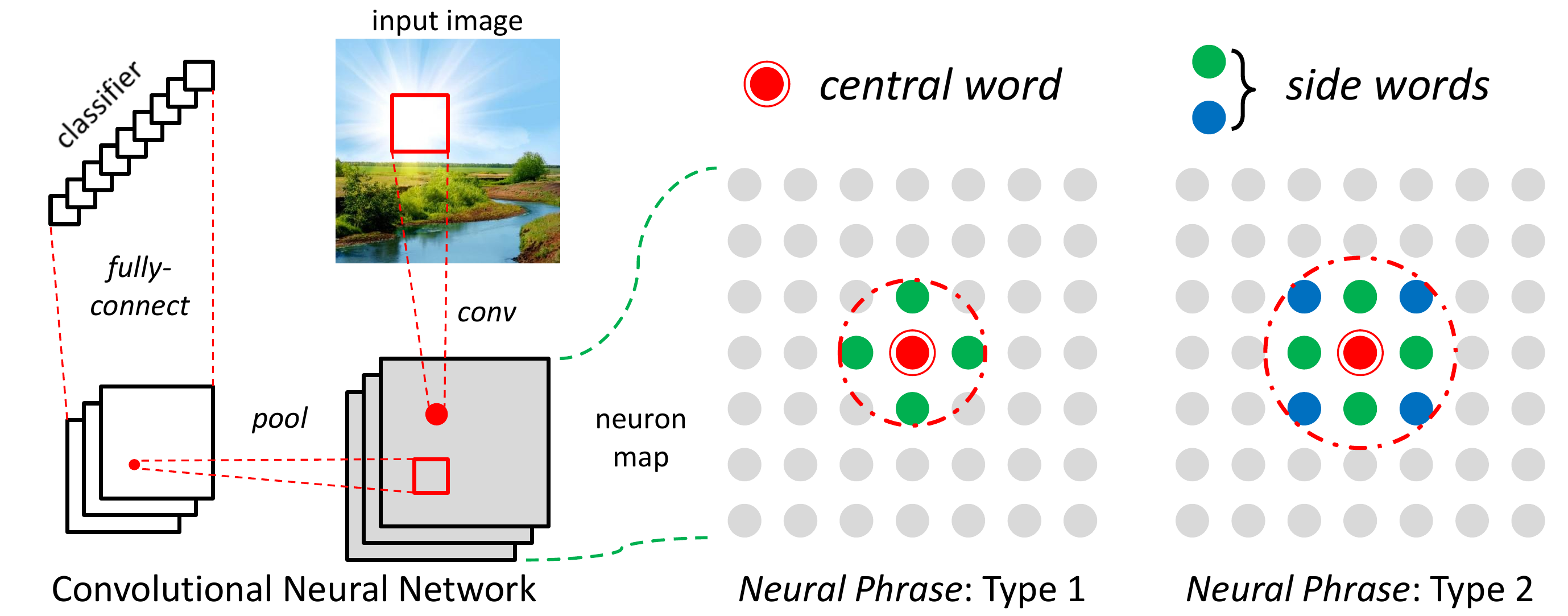}
\end{center}
\caption{
    Left: the architecture of a toy CNN model.
    A geometric neural phrase is defined on the basis of a set of neural words.
    Right: two types of neighborhood used in this work (best viewed in color PDF).
    The green side words are weighted by $\sigma$ and the blue ones by $\sigma^2$, where $\sigma$ is the smoothing parameter.
}
\label{Fig:GeometricNeuralPhrases}
\end{figure}

A {\em geometric neural phrase} is defined as
${\mathcal{G}_{w,h}}={\left\{\mathbf{x}_{w',h'}\mid{\mathbf{x}_{w',h'}}\in{\mathcal{N}_{w,h}}\right\}}$,
where $\mathcal{N}_{w,h}$ is the neighborhood of $\mathbf{x}_{w,h}$.
Given the number of side words $K$,
we can rewrite it as ${\mathcal{G}_{w,h}}={\mathbf{x}_{w,h}\cup\left\{\mathbf{x}_{w,h}^{\left(k\right)}\right\}_{k=1}^K}$,
where $\mathbf{x}_{w,h}$ is the {\em central word}, and all the others in $\mathcal{G}_{w,h}$ are {\em side words}.
For simplicity, we consider two fixed types of neighborhood, shown in Figure~\ref{Fig:GeometricNeuralPhrases}.
If the central word is located on the boundary of the neuron map, the side words outside the map are simply ignored.

The {\bf Geometric Neural Phrase Pooling} (GNPP) algorithm
computes an updated neural response for each geometric neural phrase $\mathcal{G}_{w,h}$ individually:
\begin{equation}
\label{Eqn:GNPP}
{\mathbf{z}_{w,h}}={\frac{1}{2}\left[\mathbf{x}_{w,h}+
    \max_{k=1}^K\left\{s_{w,h}^{\left(k\right)}\times\mathbf{x}_{w,h}^{\left(k\right)}\right\}\right]},
\end{equation}
where $\max_{k=1}^K\left\{\cdot\right\}$ is the maximization over $K$ side words.
Note that the central word is not included in the maximization.
We add the coefficient $\frac{1}{2}$ to approximately preserve the average scale of neuron responses.
We define a smoothing parameter ${\sigma}\in{\left(0,1\right]}$.
A side word is weighted by either ${s_{w,h}^{\left(k\right)}}={\sigma}$ or ${s_{w,h}^{\left(k\right)}}={\sigma^2}$,
according to the relative position to the central word.
Of course, one can modify the definition of both neighborhood and weights,
{\em e.g.}, using a larger neighborhood or assigning smaller weights on side words,
but these minor changes do not impact much on the performance (see Section~\ref{Experiments:Analysis}).

GNPP averages neuron responses over the central word and the maximal candidate among all side words.
Although this looks like the behavior of a local smoother,
we emphasize that GNPP is intrinsically different from other smoothers such as vanilla Gaussian blur.
Gaussian blur can be formulated as convoluting the input data with a fixed kernel.
Applying Gaussian blur after a convolutional layer is similar to using larger kernels,
where some weights are not independent to each other.
As expected, adding Gaussian blur does not obtain accuracy gain.
We add a vanilla Gaussian blur layer before each pooling layer of the {\bf LeNet}, and test it on {\bf CIFAR10}.
The baseline error rate is $17.07\%\pm0.15\%$, and the network with Gaussian blur reports $17.05\%\pm0.13\%$.
On the other side, the network with GNPP reports $14.78\%\pm0.17\%$ (see Section~\ref{Experiments:CIFAR}).
In summary, {\bf GNPP does something that a linear smoother cannot do}.

Since GNPP does not change the dimension ($W$, $H$ and $D$) of the neuron cube,
we can regard GNPP as an intermediate network layer, {\em i.e.}, the {\bf GNPP layer}.
Although the GNPP layer can, at least in theory, appear anywhere,
{\bf we only insert it between a convolutional layer and a pooling layer}, due to the reason to be elaborated in the next subsection.

\subsection{Modeling the Spatial Co-occurrence}
\label{GNPP:Explanation}

We show that GNPP is an implicit way of punishing {\em isolated} neuron responses.
Therefore, GNPP works well on the assumption that {\em clustered} neuron responses are more reliable than {\em isolated} ones.
In this subsection, we will elaborate that such an assumption is better satisfied on the high-level layers of a CNN.

\begin{figure}[t]
\begin{center}
    \includegraphics[width=\figurewidth]{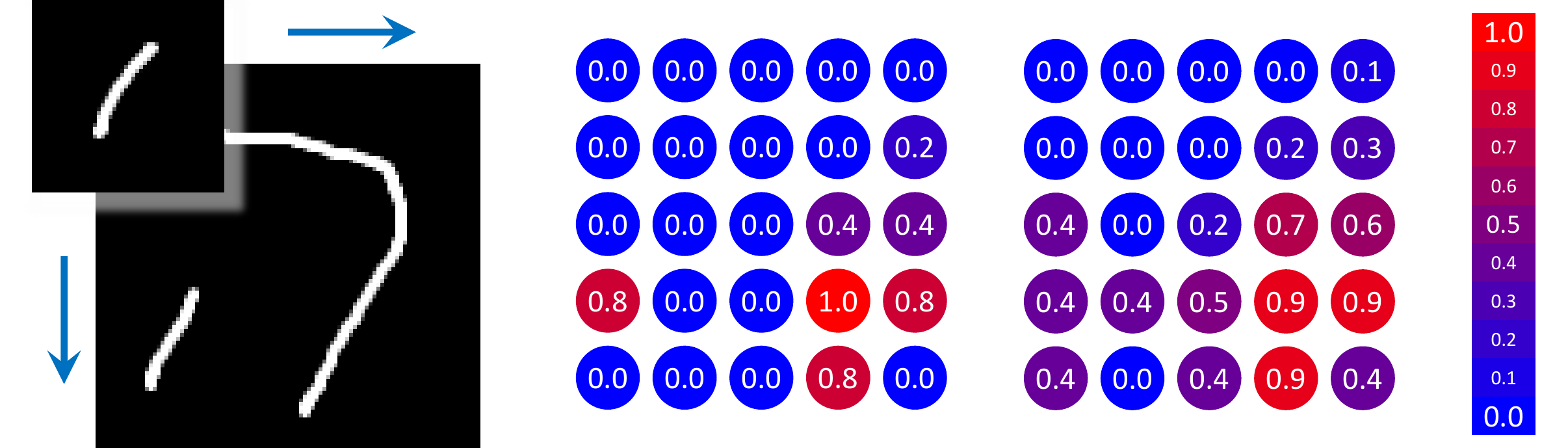}
\end{center}
\caption{
    A conceptual illustration of GNPP (best viewed in color PDF).
    Left: an image is convoluted with a template.
    Middle: the original one-dimensional neuron responses.
    Right: the responses after GNPP (type 1, ${\sigma}={1.0}$).
    The isolated high response (around the bottom-left corner) is decreased and smoothed,
    while the clustered high responses (around the bottom-right corner) are preserved.
}
\label{Fig:GNPPIllustration}
\end{figure}

To start, we note that the computation in~\eqref{Eqn:GNPP} is carried out in parallel over the $D$ channels.
Without loss of generality, we only consider a single channel in a hidden layer,
and our conclusion remains valid for the entire layer (containing $D$ channels).
In other words, we can simplify to the situation where we are dealing with $W\times H$ one-dimensional visual words.

In a CNN, neuron responses in one layer are generated by convolution.
Convoluting data with the kernel can be regarded as template matching on different spatial locations.
After ReLU activation~\cite{Krizhevsky_2012_ImageNet},
the preserved positive neuron responses correspond to those local patches with high similarity to the template.
Figure~\ref{Fig:GNPPIllustration} shows a toy example of (ReLU-activated) convolution results,
in which we can find some {\em clustered} high responses and some {\em isolated} ones.
Since GNPP averages the neuron responses over the central word and the maximal candidate of side words,
the {\em clustered} responses are approximately preserved, while the {\em isolated} ones are punished.
A toy example is shown in Figure~\ref{Fig:GNPPIllustration}.

We explain why clustered responses are more reliable, especially on a high-level layer,
where the isolated responses often correspond to unexpected random noise~\cite{Wang_2015_Discovering}.
This is because high-level convolutional kernels are highly ``specialized'',
{\em i.e.}, they often represent concrete visual concepts,
{\em e.g.}, {\em car wheel} or {\em aircraft nose}~\cite{Wang_2015_Discovering}.
Meanwhile, as the network level goes up, the receptive field of a neuron becomes larger
({\em e.g.}, a neuron on the {\em conv-5} layer of the {\bf AlexNet} can ``see'' $163\times163$ pixels on the input image),
and neighboring neurons share more and more common visual information
({\em e.g.}, the overlapping rate of two neighboring neurons on the {\em conv-5} layer is $90.2\%$).
Thus, if a positive neuron response is caused by the correct match of a visual concept,
its neighboring neurons are also likely to be activated, leading to a cluster of positive neuron responses.
Oppositely, if it is caused by some random noise,
its neighboring neurons may not be activated, and this isolated response shall be punished.

In conclusion, the core idea of GNPP is to model the spatial co-occurrence of neurons produced by a convolutional layer,
or find reliable features by punishing the isolated responses which are more likely to be unexpected random noise.
We note that pooling, when applied right after GNPP, is an efficient way of aggregating these rectified neuron responses.
Therefore, in this work, we only insert GNPP between a convolutional layer and a pooling layer.

\subsection{Comparison to Other Work}
\label{GNPP:Comparison}

The GNPP algorithm is inspired by the GPP algorithm which originates from the BoVW model (see Section~\ref{RelatedWork:GPP}).
GPP models the spatial context of visual words, and GNPP models the spatial co-occurrence of neural words.
In the BoVW model, GPP can only be applied before a max-pooling layer, but GNPP can be inserted anywhere into the CNN model.
In the {\bf SVHN} and {\bf CIFAR} experiments, we also show that GNPP cooperates well with the average-pooling layers.

The GNPP layer is related to the Spatial Pyramid Pooling (SPP) layer~\cite{He_2014_Spatial}
and the Region-Of-Interest (ROI) pooling layer~\cite{Ren_2015_Faster}.
However, the motivation and working mechanism of GNPP are quite different from these two layers.
The goal of GNPP is to punish isolated neuron responses and improve the descriptive power of every single neuron,
while the SPP layer and the ROI pooling layer aim at summarizing local neurons into a regional description.
The basic unit in the GNPP layer is a single neuron, and pooling is performed on a small set of its neighboring neurons,
whereas both the SPP layer and the ROI pooling layer work on image regions.
Finally, we point out that GNPP can be integrated with other network layers to further improve the recognition performance.

\section{Experiments}
\label{Experiments}

In this section, we show the experimental results of inserting the GNPP layer into different CNN models.
We first observe the performance by evaluating relatively shallow networks on small datasets,
then use our conclusions to inform the application of GNPP to deeper networks and the large-scale database.

\subsection{The {\bf MNIST} and {\bf SVHN} Datasets}
\label{Experiments:MNISTandSVHN}

{\bf MNIST}~\cite{LeCun_1998_Gradient} is one of the most popular datasets for handwritten digit recognition.
It contains $60\rm{,}000$ training and $10\rm{,}000$ testing samples,
uniformly distributed over $10$ categories (digits $0$--$9$).
All the samples are $28\times28$ grayscale images.
We use a modified version ($2$ convolutional layers) of the {\bf LeNet}~\cite{LeCun_1990_Handwritten} as the baseline.
With abbreviated notation, the network configuration is written as:
\begin{spverbatim}
{C5(S1P0)@20-MP2(S2)}{C5(S1P0)@50-MP2(S2)}{FC500}{FC10}.
\end{spverbatim}
\noindent
Here, {\tt C5(S1P0)@20} denotes a convolutional layer with $20$ kernels of size $5\times5$, spatial stride $1$ and padding width $0$,
{\tt MP2(S2)} is a max-pooling layer with pooling region $2\times2$ and spatial stride $2$,
and {\tt FC500} is a fully-connected layer with $500$ outputs.
All the convolution results are activated by ReLU~\cite{Krizhevsky_2012_ImageNet}, and we use the softmax loss function.
In the later experiments, we will directly use the same notations.
We apply $20$ training epochs with learning rate $10^{-3}$, followed by $4$ epochs with learning rate $10^{-4}$,
and another $1$ epoch with learning rate $10^{-5}$.
We test each network five times individually with different initialization and report the averaged error rate and standard deviation.

The {\bf SVHN} dataset~\cite{Netzer_2011_Reading} is a larger collection of $32\times32$ RGB images,
with $73\rm{,}257$ training samples, $26\rm{,}032$ testing samples, and $531\rm{,}131$ extra training samples.
We split the data as in the previous methods~\cite{Netzer_2011_Reading},
{\em i.e.}, preserving $6\rm{,}000$ images for validation, and using the remainder for training.
We use Local Contrast Normalization (LCN) for data preprocessing,
following~\cite{Sermanet_2012_Convolutional}\cite{Zeiler_2013_Stochastic}\cite{Goodfellow_2013_Maxout}.
We use another version of the {\bf LeNet} with $3$ convolutional layers, abbreviated as:
\begin{spverbatim}
{C5(S1P2)@32-MP3(S2)}{C5(S1P2)@32-AP3(S2)}{C5(S1P2)@64-AP3(S2)}{FC10}.
\end{spverbatim}
\noindent
Here, {\tt AP} indicates an average-pooling layer.
We apply $12$ training epochs with learning rate $10^{-3}$, followed by $2$ epochs with learning rate $10^{-4}$,
and another $1$ epoch with learning rate $10^{-5}$.
Each network is individually tested five times.

\newcommand{\colwidthA}{1.6cm}
\begin{table*}[t]
\begin{center}
\subfloat[\normalsize{{\bf MNIST}, with the $2$-layer {\bf LeNet}, no Dropout}]{
\begin{tabular}{|c|c||R{\colwidthA}|R{\colwidthA}|R{\colwidthA}||R{\colwidthA}|R{\colwidthA}|R{\colwidthA}|}
\hline
L$1$ & L$2$ & T1($1.0$) & T1($0.9$) & T1($0.8$) & T2($1.0$) & T2($0.9$) & T2($0.8$) \\
\hline\hline
{}         & {}         & $ 0.87\pm.02$          & $ 0.87\pm.02$          & $ 0.87\pm.02$          &
                          $ 0.87\pm.02$          & $ 0.87\pm.02$          & $ 0.87\pm.02$          \\
\hline
\checkmark & {}         & $ 0.72\pm.04$          & $ 0.73\pm.03$          & $ 0.70\pm.05$          &
                          $ 0.71\pm.06$          & $ 0.71\pm.06$          & $ 0.72\pm.04$          \\
\hline
{}         & \checkmark & $ 0.75\pm.03$          & $ 0.79\pm.02$          & $ 0.77\pm.05$          &
                          $ 0.73\pm.04$          & $ 0.75\pm.04$          & $ 0.73\pm.05$          \\
\hline
\checkmark & \checkmark & $\mathbf{ 0.72}\pm.03$ & $\mathbf{ 0.67}\pm.04$ & $\mathbf{ 0.69}\pm.04$ &
                          $\mathbf{ 0.63}\pm.03$ & $\mathbf{ 0.64}\pm.03$ & $\mathbf{ 0.67}\pm.03$ \\
\hline
\end{tabular}
} \\
\subfloat[\normalsize{{\bf MNIST}, with the $2$-layer {\bf LeNet}, Dropout ratio $0.5$}]{
\begin{tabular}{|c|c||R{\colwidthA}|R{\colwidthA}|R{\colwidthA}||R{\colwidthA}|R{\colwidthA}|R{\colwidthA}|}
\hline
L$1$ & L$2$ & T1($1.0$) & T1($0.9$) & T1($0.8$) & T2($1.0$) & T2($0.9$) & T2($0.8$) \\
\hline\hline
{}         & {}         & $ 0.72\pm.03$          & $ 0.72\pm.03$          & $ 0.72\pm.03$          &
                          $ 0.72\pm.03$          & $ 0.72\pm.03$          & $ 0.72\pm.03$          \\
\hline
\checkmark & {}         & $ 0.59\pm.02$          & $ 0.61\pm.05$          & $ 0.62\pm.03$          &
                          $ 0.59\pm.03$          & $ 0.59\pm.02$          & $ 0.63\pm.03$          \\
\hline
{}         & \checkmark & $ 0.63\pm.03$          & $ 0.62\pm.07$          & $ 0.64\pm.03$          &
                          $ 0.62\pm.05$          & $ 0.60\pm.03$          & $ 0.65\pm.03$          \\
\hline
\checkmark & \checkmark & $\mathbf{ 0.58}\pm.05$ & $\mathbf{ 0.55}\pm.02$ & $\mathbf{ 0.57}\pm.02$ &
                          $\mathbf{ 0.54}\pm.05$ & $\mathbf{ 0.56}\pm.04$ & $\mathbf{ 0.61}\pm.05$ \\
\hline
\end{tabular}
} \\
\subfloat[\normalsize{{\bf SVHN}, with the $3$-layer {\bf LeNet}, no Dropout}]{
\begin{tabular}{|c|c|c||R{\colwidthA}|R{\colwidthA}|R{\colwidthA}||R{\colwidthA}|R{\colwidthA}|R{\colwidthA}|}
\hline
L$1$ & L$2$ & L$3$ & T1($1.0$) & T1($0.9$) & T1($0.8$) & T2($1.0$) & T2($0.9$) & T2($0.8$) \\
\hline\hline
{}         & {}         & {}         & $ 4.63\pm.06$          & $ 4.63\pm.06$          & $ 4.63\pm.06$          &
                                       $ 4.63\pm.06$          & $ 4.63\pm.06$          & $ 4.63\pm.06$          \\
\hline
\checkmark & {}         & {}         & $ 4.46\pm.06$          & $ 4.47\pm.05$          & $ 4.42\pm.09$          &
                                       $ 4.42\pm.08$          & $ 4.42\pm.07$          & $ 4.43\pm.09$          \\
\hline
{}         & \checkmark & {}         & $ 4.15\pm.08$          & $ 4.18\pm.01$          & $ 4.17\pm.07$          &
                                       $ 4.08\pm.10$          & $ 4.19\pm.07$          & $ 4.20\pm.05$          \\
\hline
{}         & {}         & \checkmark & $ 3.76\pm.03$          & $ 3.72\pm.05$          & $ 3.77\pm.06$          &
                                       $ 3.53\pm.07$          & $ 3.64\pm.07$          & $ 3.65\pm.10$          \\
\hline
\checkmark & \checkmark & {}         & $ 4.10\pm.05$          & $ 4.07\pm.03$          & $ 4.10\pm.05$          &
                                       $ 4.10\pm.07$          & $ 4.10\pm.03$          & $ 4.14\pm.07$          \\
\hline
\checkmark & {}         & \checkmark & $ 3.55\pm.10$          & $ 3.60\pm.03$          & $ 3.67\pm.06$          &
                                       $ 3.47\pm.05$          & $ 3.47\pm.02$          & $ 3.55\pm.09$          \\
\hline
{}         & \checkmark & \checkmark & $\mathbf{ 3.43}\pm.06$ & $ 3.52\pm.07$          & $\mathbf{ 3.55}\pm.04$ &
                                       $\mathbf{ 3.41}\pm.03$ & $ 3.42\pm.04$          & $ 3.51\pm.05$          \\
\hline
\checkmark & \checkmark & \checkmark & $ 3.46\pm.07$          & $\mathbf{ 3.47}\pm.06$ & $\mathbf{ 3.55}\pm.06$ &
                                       $ 3.43\pm.05$          & $\mathbf{ 3.39}\pm.01$ & $\mathbf{ 3.46}\pm.03$ \\
\hline
\end{tabular}
}
\caption{
    Classification error rates ($\%$) on {\bf MNIST} and {\bf SVHN}.
    L1, L2 and L3 are three pooling layers, `\checkmark' denotes that GNPP is added.
    T1 and T2 indicate two types of neighborhood (see Figure~\ref{Fig:GeometricNeuralPhrases}).
    $1.0$, $0.9$ and $0.8$ are $\sigma$ values.
}
\label{Tab:MNISTandSVHN}
\end{center}
\end{table*}

When the GNPP layer is inserted into the network, it can appear before any subset of the pooling layers.
We enumerate all the possibilities, and summarize the results in Table~\ref{Tab:MNISTandSVHN}.
One can observe that the use of GNPP significantly improves the recognition accuracy.
The relative error rates are decreased by over $20\%$ on both datasets.
Meanwhile, GNPP can be used with Dropout~\cite{Hinton_2012_Improving}
(randomly discarding some neuron responses on the second pooling layer):
on {\bf MNIST}, the error rate is reduced from $0.72\%$ to $0.58\%$.

\subsection{The {\bf CIFAR10} and {\bf CIFAR100} Datasets}
\label{Experiments:CIFAR}

\begin{table*}[t]
\begin{center}
\subfloat[\normalsize{{\bf CIFAR10}, with the $3$-layer {\bf LeNet}, Dropout ratio $0.2$}]{
\begin{tabular}{|c|c|c||R{\colwidthA}|R{\colwidthA}|R{\colwidthA}||R{\colwidthA}|R{\colwidthA}|R{\colwidthA}|}
\hline
L$1$ & L$2$ & L$3$ & T1($1.0$) & T1($0.9$) & T1($0.8$) & T2($1.0$) & T2($0.9$) & T2($0.8$) \\
\hline\hline
{}         & {}         & {}         & $17.07\pm.15$          & $17.07\pm.15$          & $17.07\pm.15$          &
                                       $17.07\pm.15$          & $17.07\pm.15$          & $17.07\pm.15$          \\
\hline
\checkmark & {}         & {}         & $16.67\pm.22$          & $16.80\pm.25$          & $16.84\pm.12$          &
                                       $16.65\pm.19$          & $17.03\pm.15$          & $17.04\pm.17$          \\
\hline
{}         & \checkmark & {}         & $15.79\pm.22$          & $16.09\pm.17$          & $15.95\pm.31$          &
                                       $15.69\pm.11$          & $16.07\pm.27$          & $15.90\pm.09$          \\
\hline
{}         & {}         & \checkmark & $15.49\pm.15$          & $15.31\pm.20$          & $15.51\pm.25$          &
                                       $15.27\pm.10$          & $15.29\pm.14$          & $15.28\pm.16$          \\
\hline
\checkmark & \checkmark & {}         & $15.82\pm.23$          & $15.76\pm.18$          & $15.98\pm.14$          &
                                       $16.05\pm.29$          & $15.90\pm.25$          & $15.94\pm.09$          \\
\hline
\checkmark & {}         & \checkmark & $15.15\pm.20$          & $15.29\pm.12$          & $15.44\pm.19$          &
                                       $15.29\pm.32$          & $15.19\pm.35$          & $15.20\pm.35$          \\
\hline
{}         & \checkmark & \checkmark & $\mathbf{14.92}\pm.18$ & $15.00\pm.18$          & $15.15\pm.15$          &
                                       $\mathbf{14.83}\pm.25$ & $14.93\pm.20$          & $14.92\pm.16$          \\
\hline
\checkmark & \checkmark & \checkmark & $14.97\pm.17$          & $\mathbf{14.83}\pm.23$ & $\mathbf{14.78}\pm.17$ &
                                       $15.22\pm.16$          & $\mathbf{14.79}\pm.26$ & $\mathbf{14.85}\pm.26$ \\
\hline
\end{tabular}
} \\
\subfloat[\normalsize{{\bf CIFAR100}, with the $3$-layer {\bf LeNet}, no Dropout}]{
\begin{tabular}{|c|c|c||R{\colwidthA}|R{\colwidthA}|R{\colwidthA}||R{\colwidthA}|R{\colwidthA}|R{\colwidthA}|}
\hline
L$1$ & L$2$ & L$3$ & T1($1.0$) & T1($0.9$) & T1($0.8$) & T2($1.0$) & T2($0.9$) & T2($0.8$) \\
\hline\hline
{}         & {}         & {}         & $44.99\pm.19$          & $44.99\pm.19$          & $44.99\pm.19$          &
                                       $44.99\pm.19$          & $44.99\pm.19$          & $44.99\pm.19$          \\
\hline
\checkmark & {}         & {}         & $44.62\pm.17$          & $44.53\pm.45$          & $44.78\pm.06$          &
                                       $44.43\pm.29$          & $44.58\pm.36$          & $44.58\pm.52$          \\
\hline
{}         & \checkmark & {}         & $43.34\pm.23$          & $43.71\pm.19$          & $43.37\pm.26$          &
                                       $43.21\pm.23$          & $43.03\pm.27$          & $43.37\pm.30$          \\
\hline
{}         & {}         & \checkmark & $43.11\pm.24$          & $42.77\pm.37$          & $42.99\pm.24$          &
                                       $42.96\pm.32$          & $42.81\pm.38$          & $43.08\pm.39$          \\
\hline
\checkmark & \checkmark & {}         & $43.99\pm.07$          & $43.63\pm.11$          & $43.50\pm.26$          &
                                       $43.38\pm.37$          & $43.34\pm.27$          & $43.46\pm.25$          \\
\hline
\checkmark & {}         & \checkmark & $42.85\pm.38$          & $42.81\pm.27$          & $42.82\pm.29$          &
                                       $43.08\pm.27$          & $42.79\pm.34$          & $42.93\pm.22$          \\
\hline
{}         & \checkmark & \checkmark & $\mathbf{42.35}\pm.30$ & $\mathbf{42.34}\pm.31$ & $\mathbf{42.04}\pm.20$ &
                                       $\mathbf{42.92}\pm.33$ & $\mathbf{42.72}\pm.25$ & $\mathbf{42.54}\pm.29$ \\
\hline
\checkmark & \checkmark & \checkmark & $42.97\pm.29$          & $42.77\pm.36$          & $42.36\pm.18$          &
                                       $43.31\pm.34$          & $42.85\pm.18$          & $42.60\pm.36$          \\

\hline
\end{tabular}
}
\caption{
    Classification error rates ($\%$) on the {\bf CIFAR} datasets.
    We use the same notations as in Table~\ref{Tab:MNISTandSVHN}.
    We apply Dropout on the simpler {\bf CIFAR10} task.
}
\label{Tab:CIFAR}
\end{center}
\end{table*}

Both {\bf CIFAR10} and {\bf CIFAR100} datasets~\cite{Krizhevsky_2009_Learning}
are subsets of the $80$-million tiny image database~\cite{Torralba_2008_80}.
Both of them have $50\rm{,}000$ training and $10\rm{,}000$ testing samples,
uniformly distributed over $10$ or $100$ categories.
We also use the $3$-layer {\bf LeNet} as in {\bf SVHN}, with the fully-connected layer replaced by {\tt FC100} in {\bf CIFAR100}.
We augment the training data by randomly flipping each training image with $50\%$ probability.
We apply $120$ training epochs with learning rate $10^{-3}$, followed by $20$ epochs with learning rate $10^{-4}$,
and another $10$ epochs with learning rate $10^{-5}$.
Each network is individually tested five times.

Results with all possible GNPP settings are summarized in Table~\ref{Tab:CIFAR}.
Once again, GNPP improves the baseline error rate significantly:
the baseline error rates on both {\bf CIFAR10} and {\bf CIFAR100} are reduced by more than $2\%$,
and the relative error rate decrease are $11.25\%$ and $6.56\%$, respectively.

\subsection{Analysis on Small Experiments}
\label{Experiments:Analysis}

Before we go into deeper networks and larger datasets, we conduct some preliminary analysis based on the results we already have.

First, although inserting GNPP before any pooling layers improves the performance,
the most significant accuracy gain brought by a single GNPP layer is obtained by adding GNPP before the last pooling layer.
This reinforces the conclusion drawn in Section~\ref{GNPP:Explanation},
{\em i.e.}, GNPP works better on the high-level neuron responses.
Meanwhile, on the {\bf SVHN} and {\bf CIFAR} datasets,
adding GNPP before all three pooling layers produces inferior results to that adding GNPP before the second and third pooling layers.
In the later experiments, we first add the GNPP layer before each pooling layer individually,
then use the results to inform the design of the final model.

Regarding the scale of neural phrases, {\em i.e.}, $K$,
we find that increasing the scale is not guaranteed to produce better recognition results.
We explain this by noticing that adding a faraway side word to a neural phrase, most often,
does not provide much related information but risks introducing noise to that unit.
This idea can also be used to explain why a proper smoothing parameter, say, ${\sigma}={0.8}$,
helps to reduce the contribution of faraway side words, leading to better recognition performance.
One may certainly try other choices such as a large neighborhood with a very small $\sigma$,
but we note that the time complexity of a GNPP layer is linear to $K$.
In the later experiments, we will directly use the first type of neighborhood (${K}={4}$) with ${\sigma}={0.8}$.

\subsection{Deeper Networks and the State-of-the-Arts}
\label{Experiments:DeeperNetworks}

We adopt two deeper networks on the above four small datasets to compare with the state-of-the-art results.
One of them (we name it as the {\bf BigNet}) is borrowed from~\cite{Nagadomi_2014_Kaggle} in the Kaggle recognition competition,
and other one one is the $16$-layer Wide Residual Network ({\bf WRN})~\cite{Zagoruyko_2016_Wide} with dropout.
Both networks can be used in each of the four small datasets.
In {\bf CIFAR} datasets, we randomly flip the image with $50\%$ probability.
We train the {\bf BigNet} using $6\times10^6$ samples with learning rate $10^{-2}$,
followed by $3\times10^6$ samples with learning rate $10^{-3}$ and $1\times10^6$ samples with learning rate $10^{-4}$, respectively.
We report a $7.80\%$ error rate on {\bf CIFAR10}, comparable to the original version~\cite{Nagadomi_2014_Kaggle},
which uses a very complicated way of data preparation and augmentation to get a $6.68\%$ error rate.
Training our model needs about $1$ hour, while the original version~\cite{Nagadomi_2014_Kaggle} requires $6$ hours.
We train the {\bf WRN} following the original configuration in~\cite{Zagoruyko_2016_Wide}.

\newcommand{\colwidthB}{1.8cm}
\begin{table*}[t]
\begin{center}
\begin{tabular}{|l||R{\colwidthB}|R{\colwidthB}|R{\colwidthB}|R{\colwidthB}|}
\hline
{} & {\bf MNIST} & {\bf SVHN} & {\bf CIFAR10} & {\bf CIFAR100} \\
\hline\hline
Zeiler {\em et.al}~\cite{Zeiler_2013_Stochastic}     & $ 0.47$          & $ 2.80$          & $15.13$          & $42.51$          \\
\hline
Goodfellow {\em et.al}~\cite{Goodfellow_2013_Maxout} & $ 0.45$          & $ 2.47$          & $ 9.38$          & $38.57$          \\
\hline
Lin {\em et.al}~\cite{Lin_2014_Network}              & $ 0.47$          & $ 2.35$          & $ 8.81$          & $35.68$          \\
\hline
Lee {\em et.al}~\cite{Lee_2015_Deeply}               & $ 0.39$          & $ 1.92$          & $ 7.97$          & $34.57$          \\
\hline
Liang {\em et.al}~\cite{Liang_2015_Recurrent}        & $\mathbf{ 0.31}$ & $ 1.77$          & $ 7.09$          & $31.75$          \\
\hline
Lee {\em et.al}~\cite{Lee_2016_Generalizing}         & $\mathbf{ 0.31}$ & $ 1.69$          & $ 6.05$          & $32.37$          \\
\hline\hline
{\bf BigNet} (without GNPP)                          & $ 0.36$          & $ 2.14$          & $ 7.80$          & $31.03$          \\
\hline
{\bf BigNet} (with GNPP)                             & $\mathbf{ 0.32}$ & $\mathbf{ 1.87}$ & $\mathbf{ 7.14}$ & $\mathbf{29.74}$ \\
\hline
{\bf WRN} (without GNPP)                             & $ 0.34$          & $ 1.77$          & $ 5.54$          & $25.52$          \\
\hline
{\bf WRN} (with GNPP)                                & $\mathbf{ 0.31}$ & $\mathbf{ 1.67}$ & $\mathbf{ 5.31}$ & $\mathbf{25.01}$ \\
\hline
\end{tabular}
\caption{
    Comparison of the recognition error rate ($\%$) with the state-of-the-arts.
    We apply data augmentation on all these datasets, but the competitors do not use it in {\bf CIFAR100}.
    Without data augmentation, we report $29.92\%$ and $29.17\%$ error rates (using {\bf WRN}) without and with GNPP, respectively.
}
\label{Tab:Comparison}
\end{center}
\end{table*}

We compare our results with the state-of-the-arts in Table~\ref{Tab:Comparison}.
We add GNPP before the second and the third pooling layers for {\bf BigNet}, and the last pooling layer for {\bf WRN}.
Although the baseline is already pretty high, GNPP still improves it by a margin:
on {\bf BigNet}, the relative error rate drops are $11.11\%$, $12.62\%$, $8.46\%$ and $4.16\%$ on the four datasets, respectively.
Without complicated tricks, our results are very competitive among these recent works.
We believe that GNPP can also be applied to other powerful networks in the future.

\subsection{ImageNet Experiments}
\label{Experiments:ImageNet}

Finally, we evaluate our model on the {\bf ImageNet} large-scale visual recognition task
(the {\bf ILSVRC2012} dataset~\cite{Russakovsky_2015_ImageNet} with $1000$ categories).
We use the {\bf AlexNet} (provided by the {\bf CAFFE} library~\cite{Jia_2014_CAFFE}), abbreviated as:
\begin{spverbatim}
{C11(S4)@96-MP3(S2)}{C5(S1P2)@256-MP3(S2)}{C3(S1P1)@384}{C3(S1P1)@384}
{C3(S1P1)@256-MP3(S2)}{FC4096-D0.5}{FC4096-D0.5}{FC1000}.
\end{spverbatim}
\noindent
The input image is of size $227\times227$, randomly cropped from the original $256\times256$ image.
Following the setting of {\bf CAFFE}, a total of $450\rm{,}000$ mini-batches (approximately $90$ epochs) are used for training,
each of which has $256$ image samples, with the initial learning rate $10^{-2}$, momentum $0.9$ and weight decay $5\times10^{-4}$.
The learning rate is decreased to $1/10$ after every $100\rm{,}000$ mini-batches.

{\bf AlexNet} contains three max-pooling layers, {\em i.e.}, {\em pool-1}, {\em pool-2} and {\em pool-5}.
After individual tests, we only add GNPP before the last one ({\em pool-5}),
since adding GNPP before either {\em pool-1} or {\em pool-2} causes accuracy drop.
With the GNPP layer, the top-$1$ and top-$5$ recognition error rates are $42.16\%$ and $19.24\%$, respectively.
Comparing to the original rates ($43.19\%$ and $19.87\%$), GNPP boosts them by about $1.0\%$ and $0.6\%$, respectively.
We emphasize that the accuracy gain is not so small as it seems,
especially considering that we do not introduce extra parameters and that the overall training time is only increased by $1.29\%$.

Although GNPP is tested on {\bf AlexNet}, we believe it can be applied to other models,
such as {\bf VGGNet}~\cite{Simonyan_2015_Very}, {\bf GoogleNet}~\cite{Szegedy_2015_Going} and Deep Residual Nets~\cite{He_2015_Deep}.

\section{Benefits of GNPP}
\label{Benefits}

This section presents several discussions and diagnostic experiments
that help us understand the side benefits brought by the GNPP layer.

\subsection{Improving Internal Representation}
\label{Benefits:Representation}

\begin{figure}[t]
\begin{center}
    \includegraphics[width=\figurewidth]{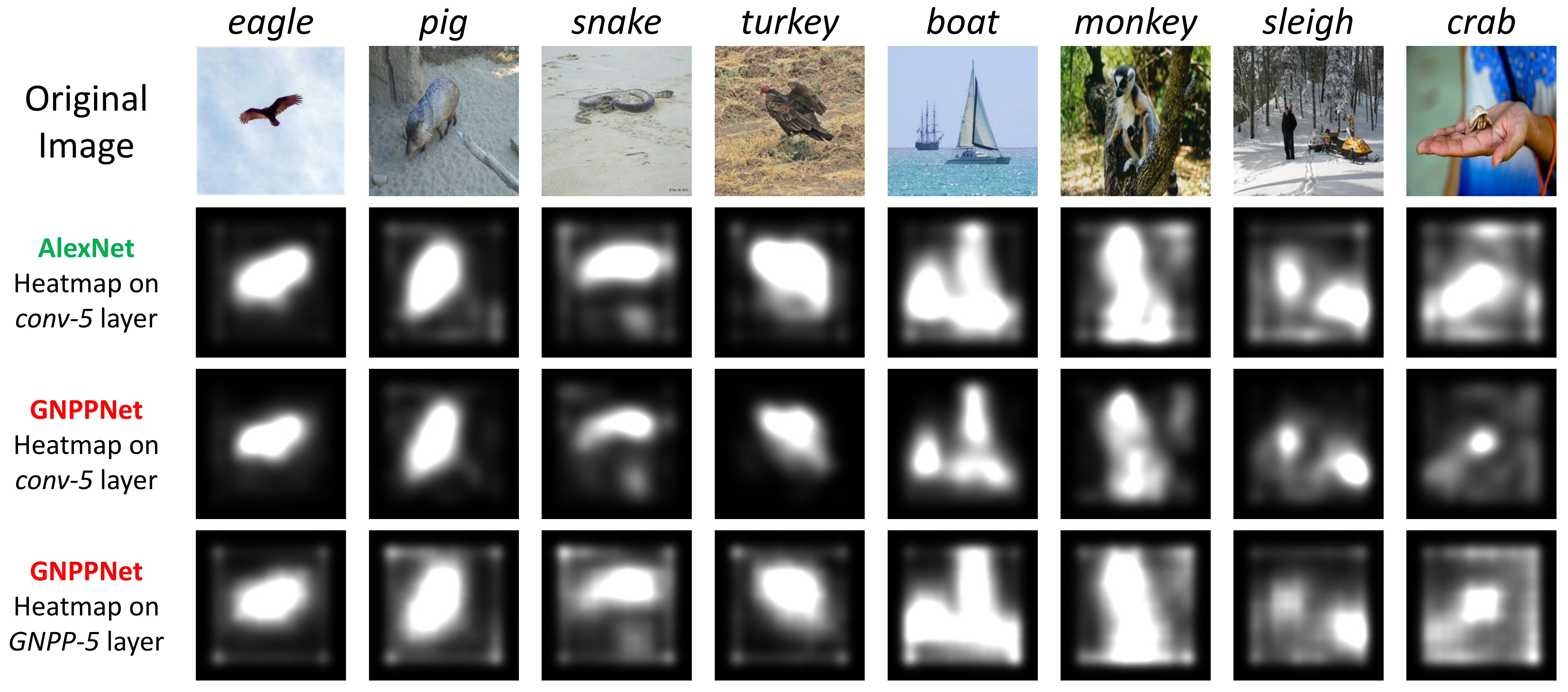}
\end{center}
\caption{
    Neuron response heatmaps produced by {\bf AlexNet} and {\bf GNPPNet}.
    When the background is relatively simple ({\em e.g.}, first two images), both methods work well.
    On those challenging cases, GNPP produces better saliency detection results,
    implying that the internal representation of CNN is improved.
}
\label{Fig:NeuronHeatmaps}
\end{figure}

Here we compare the {\em conv-5} layer of the standard {\bf AlexNet} with the corresponding layers in the {\bf GNPPNet}
(defined in Section~\ref{Experiments:ImageNet}).
That is, we compare {\bf AlexNet}'s {\em conv-5} layer with {\bf GNPPNet}'s {\em conv-5} layer and {\em GNPP-5} layer.
Each layer is a $13\times13\times256$ neuron blob corresponding to $256$ convolutional kernels.
We average over the $256$ channels and obtain a $13\times13$ heatmap.
To allow direct comparison with the input image ($227\times227$),
we diffuse each neuron response as a Gaussian distribution over its receptive field on the input image
(the same standard deviation is used on all layers).
Results are shown in Figure~\ref{Fig:NeuronHeatmaps}.
It is observed in~\cite{Wang_2015_Discovering} that
the activation patterns in higher convolutional layers correspond to mid-level parts.
The average over filters is a crude measure that some mid-level parts are detected.
Then Figure~\ref{Fig:NeuronHeatmaps} shows the spatial pattern corresponding to mid-level part detection.

First note that {\bf AlexNet}'s {\em conv-5} layer and {\bf GNPPNet}'s {\em GNPP-5} layer are broadly similar.
This is to be expected as both of them occur at corresponding places in the network architecture,
{\em i.e.}, just before the {\em pool-5} layer and the fully-connected layers.
We might think of the filter averages shown in Figure~\ref{Fig:NeuronHeatmaps}
as spatial summaries of average scores over object parts.
The higher layers in both networks combine spatial co-occurrences of parts into whole object detectors.
For example, {\em car wheels} and {\em car doors} are combined into a whole {\em car}.

Next notice that {\bf GNPPNet}'s {\em conv-5} layer is sparser and more concentrated than {\bf AlexNet}'s {\em conv-5} layer.
Broadly speaking the GNPP operation acts as a smoother
and it is the smoothed {\em conv-5} layer ({\em i.e.}, the {\em GNPP-5} layer) that resembles {\bf AlexNet}'s {\em conv-5} layer.
The difference between {\bf AlexNet}'s {\em conv-5} layer and {\bf GNPPNet}'s {\em GNPP-5} layer is subtle,
but we see that the {\em GNPP-5} layer is more diffuse corresponding to GNPP's action as local smoother.

As a result,
{\bf GNPPNet}'s {\em conv-5} layer produces better saliency detection results compared to {\bf AlexNet}'s {\em conv-5} layer.
This property can be used to extract better {\em deep features}.
We verify our hypothesis on the {\bf Caltech256} dataset~\cite{Griffin_2007_Caltech}.
$256$-dimensional feature vectors are extracted from the {\em conv-5} layer by averaging over $13\times13$ spatial locations.
The classification accuracy using the {\bf AlexNet} is $59.36\%$, and {\bf GNPPNet} improves it to $60.56\%$.
This improvement is significant given that no extra time or memory is required for feature extraction.

In summary, applying GNPP to CNN produces better internal representation.
The deep features extracted from the {\bf GNPPNet} can also benefit other vision applications,
such as image retrieval~\cite{Razavian_2014_CNN} and object detection~\cite{Girshick_2014_Rich}\cite{Girshick_2015_Fast}.

\subsection{Building Latent Connections}
\label{Benefits:LatentConnections}

We show that GNPP builds latent connections between hidden layers in the CNN model.
Consider a geometric neural phrase
${\mathcal{G}_{w,h}}={\mathbf{x}_{w,h}\cup\left\{\mathbf{x}_{w,h}^{\left(k\right)}\right\}}_{k=1}^K$.
Let $\mathcal{S}_{w,h}$ be the set of neurons in the previous layer that are connected to $\mathbf{x}_{w,h}$,
and $\mathcal{S}_{w,h}^{\left(k\right)}$ be the set connected to $\mathbf{x}_{w,h}^{\left(k\right)}$, ${k}={1,2,\ldots,K}$.
If we consider $\mathcal{G}_{w,h}$ as a {\em GNPP neuron},
then the set of neurons in the previous layer that are connected to it is
$\mathcal{S}_{w,h}\cup{\bigcup_{k=1}^{K}}\mathcal{S}_{w,h}^{\left(k\right)}$.
Thus, we are actually building latent neuron connections which do not exist in the original network.
For example, applying GNPP (type 1) before the {\em pool-5} layer of the {\bf AlexNet}
increases the number of neuron connections between {\em conv-4} and {\em conv-5}
from $149.5\mathrm{M}$ (million) to $348.9\mathrm{M}$
(on each neuron in {\em conv-5}, the number of connections to the previous layer increases from $9$ to $21$),
meanwhile the number of learnable parameters remains unchanged.

To verify the benefits of latent connections, we train another version of {\bf AlexNet}, referred to as {\bf AlexNet2},
with the difference that the number of channels on the {\em conv-5} layer increases from $256$ to $512$.
The number of neuron connections between {\em conv-4} and {\em conv-5} increases from $149.5\mathrm{M}$ to $299.0\mathrm{M}$,
comparable to $348.9\mathrm{M}$ in {\bf GNPPNet}.
{\bf AlexNet2} requires $9.97\%$ extra training time and $5.58\%$ extra GNPP memory,
while the numbers for {\bf GNPPNet} are $1.29\%$ and $2.52\%$, respectively.
{\bf AlexNet2} produces $42.45\%$ (top-$5$) and $19.47\%$ (top-$1$) recognition error rates,
which are higher than $43.19\%$ and $19.97\%$ reported by {\bf AlexNet},
but lower than $42.16\%$ and $19.24\%$ reported by {\bf GNPPNet}.
To summarize, GNPP allows latent connections to be built in an efficient manner.

\subsection{Accelerating Network Training}
\label{Benefits:Acceleration}

We show that adding GNPP layers accelerates the network training process,
since GNPP allows visual information to propagate faster, like~\cite{Srivastava_2015_Training}.

Let us investigate the case that training the $3$-layer {\bf LeNet} on the {\bf SVHN} and {\bf CIFAR} datasets.
We are interested in the following question: if the input is a $32\times32$ image,
which is the earliest layer containing a neuron able to ``see'' the entire image?
Without GNPP, we need to wait until the {\em conv-3} layer.
When GNPP is inserted before the second pooling layer, the receptive field of the neurons on the subsequent layers are increased.
Consequently, some neurons in {\em pool-2} can already ``see'' the entire image.
This allows some low-level and mid-level information ({\em e.g.}, object parts) be combined earlier.

\newcommand{\subscatterwidth}{5.5cm}
\newcommand{\subscatterkern}{0.2cm}
\begin{figure}
\begin{center}
    \begin{minipage}{\subscatterwidth}
        \centering
            \includegraphics[width=\subscatterwidth]{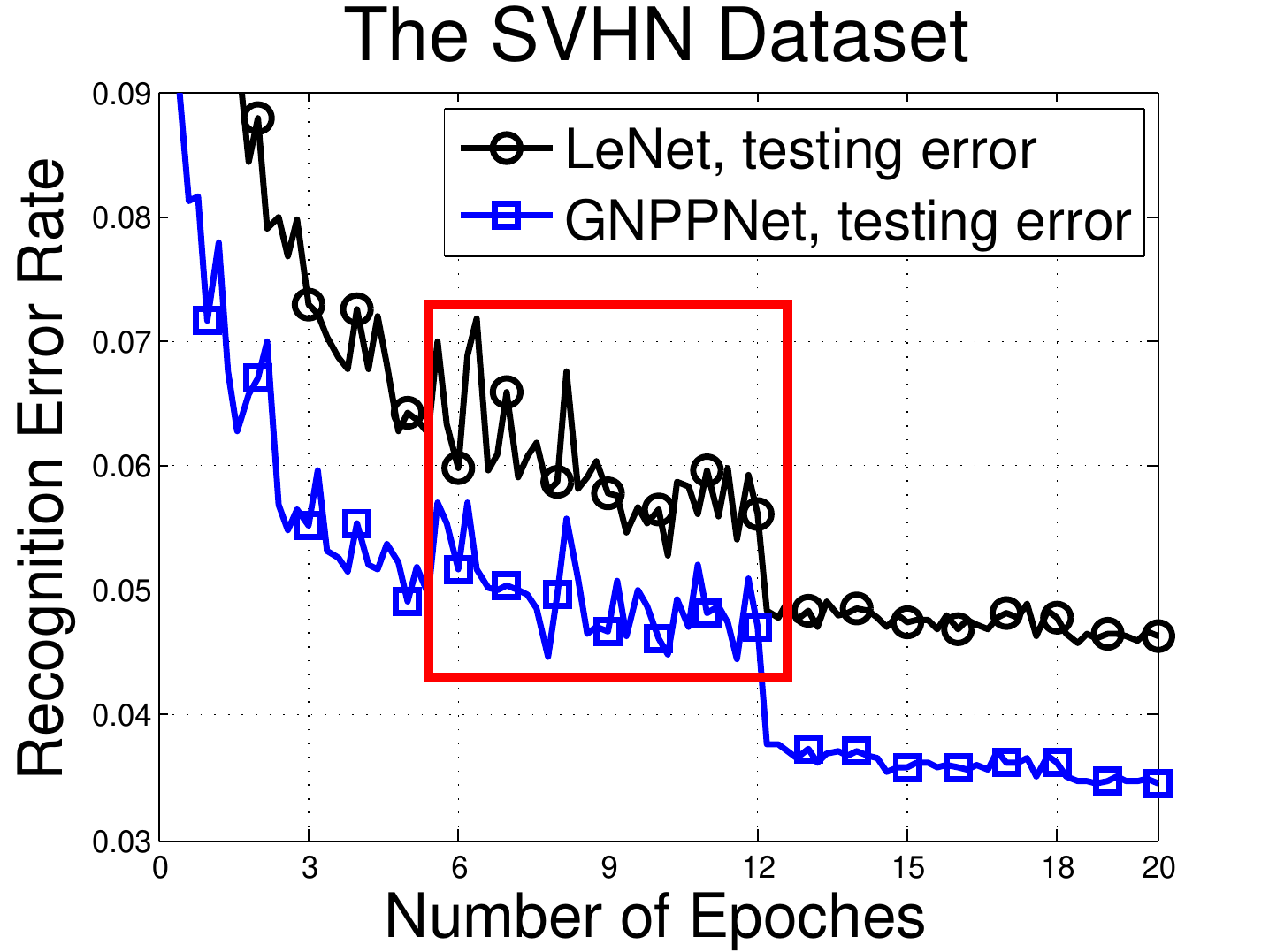}
    \end{minipage}
    \hspace{\subscatterkern}
    \begin{minipage}{\subscatterwidth}
        \centering
            \includegraphics[width=\subscatterwidth]{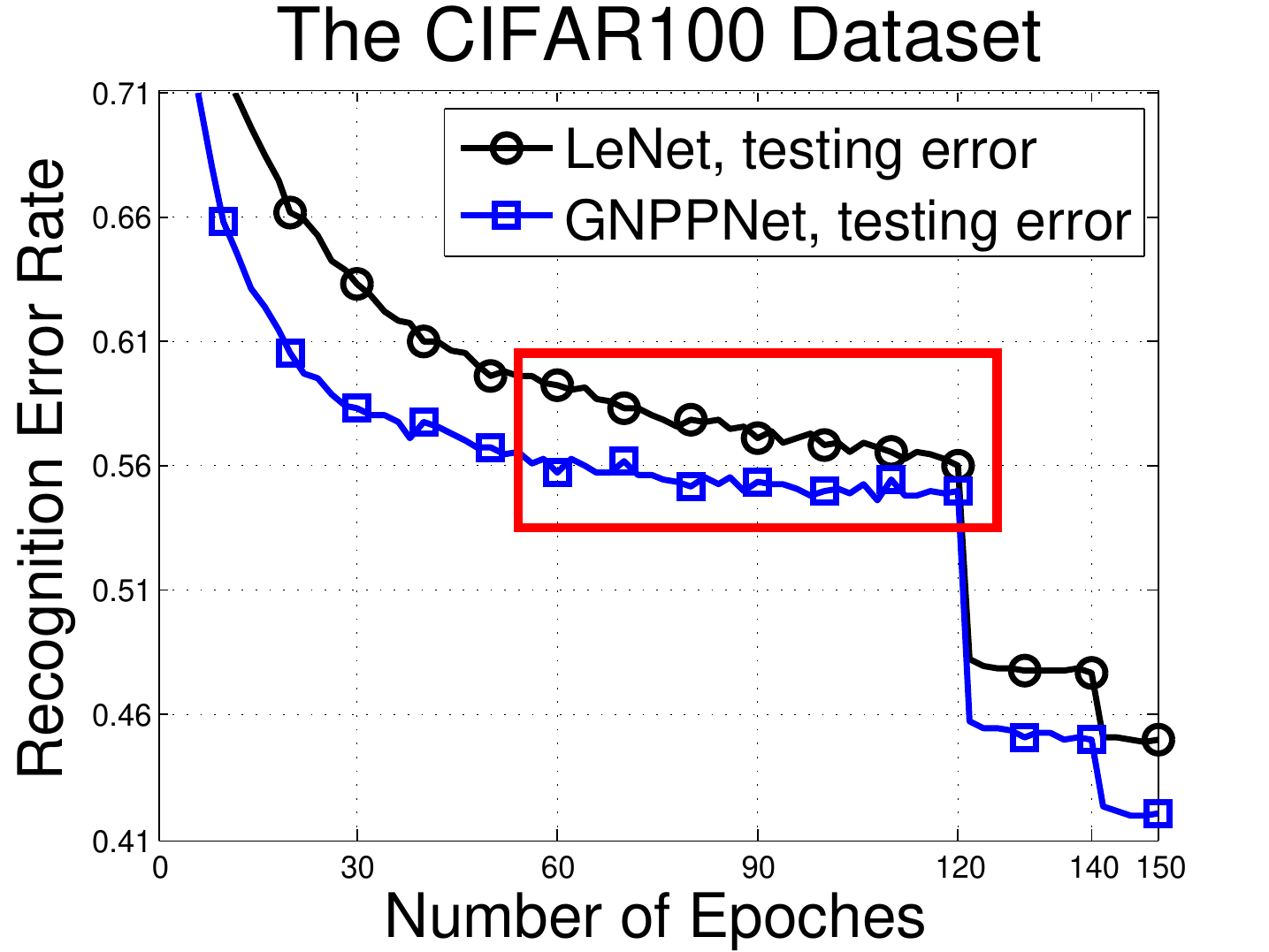}
    \end{minipage}
    \vspace{\subscatterkern}
    \vspace{\subscatterkern}
    \begin{minipage}{\subscatterwidth}
        \centering
            \includegraphics[width=\subscatterwidth]{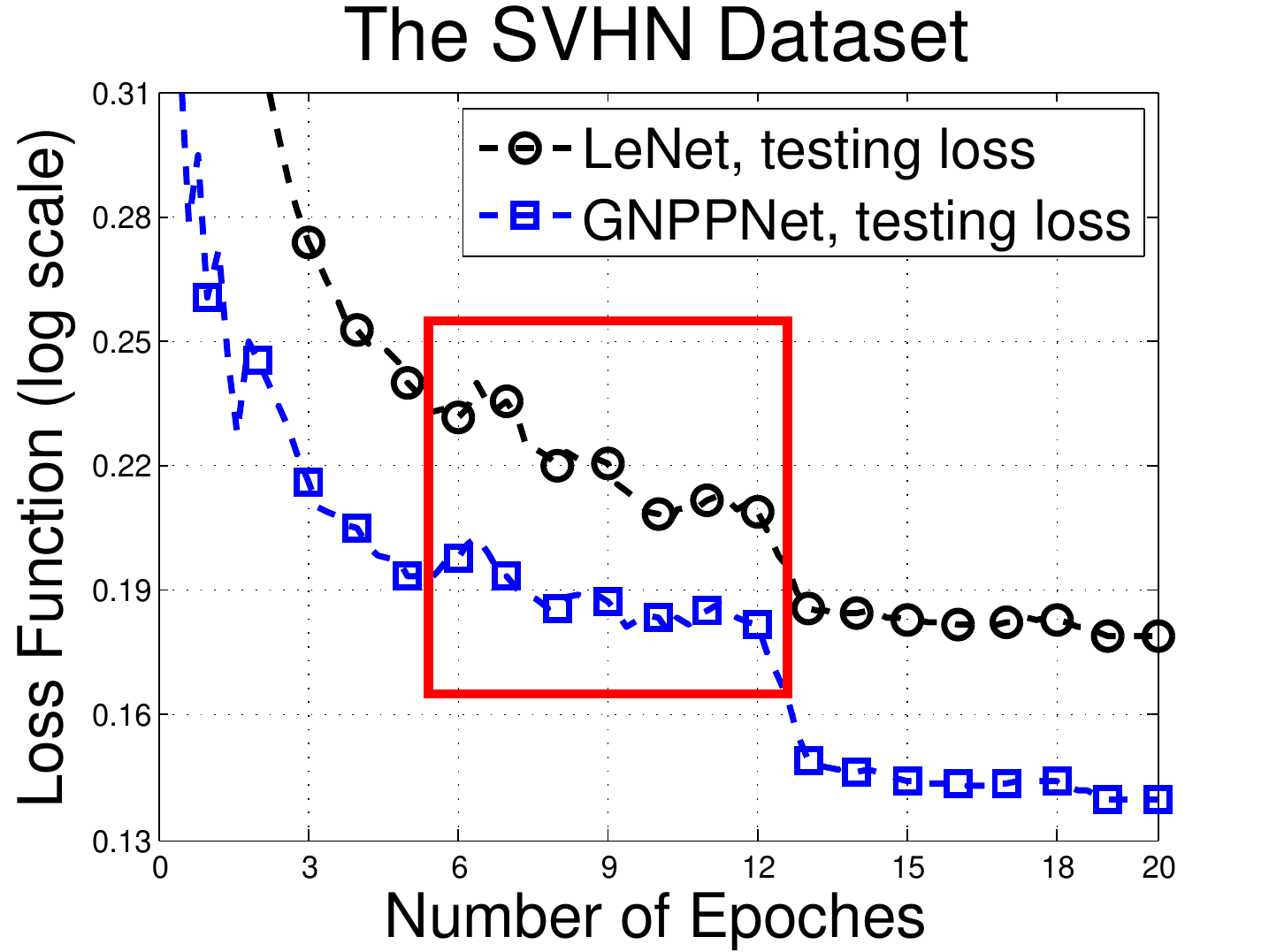}
    \end{minipage}
    \hspace{\subscatterkern}
    \begin{minipage}{\subscatterwidth}
        \centering
            \includegraphics[width=\subscatterwidth]{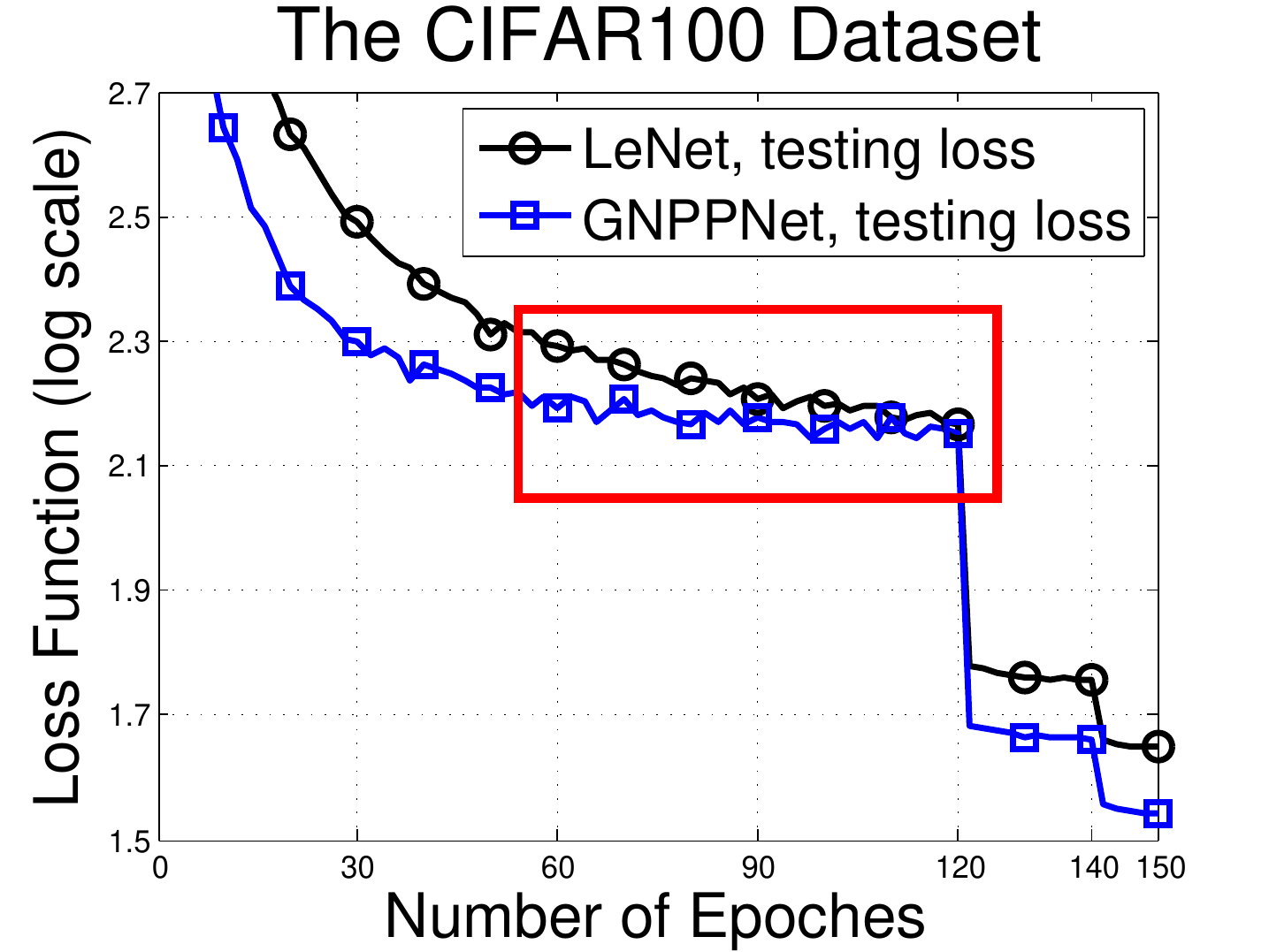}
    \end{minipage}
\end{center}
\caption{
    Error rate and loss function curves on the {\bf SVHN} and {\bf CIFAR100} datasets.
    {\bf GNPPNet} refers to the {\bf LeNet} with two GNPP layers inserted before the second and third pooling layers.
    The curves in red frames indicate that {\bf GNPPNet} enjoys better convergence, {\em i.e.}, it reaches the plateau sooner.
}
\label{Fig:Curves}
\end{figure}

As a result, GNPP helps the network training process converge faster.
To verify, we plot the testing error rates and the loss function values throughout the training process.
The results on the {\bf SVHN} and {\bf CIFAR100} datasets, using the {\bf LeNet}, are shown in Figure~\ref{Fig:Curves}.
One can see that GNPP causes the error rate and loss function curves drop more quickly, especially in the early epochs.
For example, in the {\bf SVHN} dataset, the network without GNPP requires about $36\rm{,}000$ iterations to reach $6\%$ error rate,
while that with GNPP only needs about $15\rm{,}000$ iterations to get the same rate.
Meanwhile, the training process reaches plateau sooner in the GNPP-equipped networks
(see the error rate curve between $6$--$12$ epochs in {\bf SVHN}, and that between $60$--$120$ epochs in {\bf CIFAR100}).

With the help of GNPP, we can even train a network faster and obtain better performance.
The baseline error rates on {\bf SVHN} and {\bf CIFAR100}, using the {\bf LeNet}, are $4.63\%$ and $44.99\%$, respectively.
We train a GNPP-equipped {\bf LeNet} with half training epochs under each learning rate,
and obtain $3.78\%$ and $43.35\%$ error rates (the full training reports $3.55\%$ and $42.04\%$).

\section{Conclusions}
\label{Conclusions}

In this paper, we demonstrate that constructing and encoding {\em neural phrases} boost the performance of state-of-the-art CNNs.
We insert Geometric Neural Phrase Pooling (GNPP) as an intermediate layer into the network,
and show that it improves the performance of deep networks without requiring much more computational resources.
GNPP can be explained as an implicit way of modeling the spatial co-occurrence of neurons.
We also show that GNPP enjoys the advantage of improving the internal representation of CNN,
building latent connections, and speeding up the network training process.

Our work illustrates that designing deep networks benefits from prior knowledge.
In the case of GNPP, we learn that the isolated neuron responses are less reliable than the clustered ones.
We hope that other prior knowledge can be useful incorporated into the CNN architecture.
Meanwhile, other visual tasks, including detection, segmentation, {\em etc.}, may also benefit from the GNPP algorithm.
The exploration of these topics is left for future works.

\section*{Acknowledgements}
\label{Acknowledgements}

This paper is supported in part to Prof. Alan Yuille by iARPA MICrONS contract D16PC00007 and ONR N00014-12-1-0883.
This work is supported in part to Prof. Qi Tian by ARO grants W911NF-15-1-0290
and Faculty Research Gift Awards by NEC Laboratories of America and Blippar.
This work is also supported in part by National Science Foundation of China (NSFC) 61429201.
We thank Junhua Mao, Cihang Xie and Zhuotun Zhu for insightful discussions.

\bibliographystyle{splncs}
\bibliography{egbib}
\end{document}